\documentclass[runningheads]{llncs}
\usepackage[T1]{fontenc}

\usepackage[table]{xcolor}
\usepackage{cite}
\usepackage{textcomp}
\usepackage{stfloats}
\usepackage{url}
\usepackage{verbatim}
\usepackage{graphicx}
\usepackage{times}
\usepackage{paralist}
\usepackage{amssymb}
\usepackage{mathtools}
\usepackage{tikz}
\usepackage{lipsum}
\usepackage{amsmath}
\usepackage{amssymb}
\usepackage{scalerel}
\usepackage{booktabs}

\usetikzlibrary{arrows}
\usetikzlibrary{shapes}
\usetikzlibrary{snakes}
\usetikzlibrary{calc}
\usetikzlibrary{intersections}
\usetikzlibrary{arrows}
\usetikzlibrary{shapes}
\usetikzlibrary{snakes}
\usetikzlibrary{calc}
\usetikzlibrary{tikzmark,fit}
\usetikzlibrary{positioning}
\usetikzlibrary{spy}
\usepackage{pgfplots}
\usetikzlibrary{pgfplots.statistics, pgfplots.colorbrewer, pgfplots.groupplots, pgfplots.fillbetween}
\usepackage{pgfplotstable}
\usepackage{siunitx}
\usepackage{authblk}
\usepackage{multirow}
\usepackage[hidelinks]{hyperref}

\usepackage{algorithm,algpseudocode}

\usepackage[font=footnotesize]{subcaption}
\newcommand{\citep}[1]{\cite{#1}}

\newcommand{\ulm}[1]{_{\scaleto{\mathrm{#1}}{3pt}}}

\newcommand{\param}{\theta}
\newcommand{\paramnom}{\param\ulm{nom}}

\newcommand{\sysstate}[1]{s(#1)}
\newcommand{\sysinput}[1]{a(#1)}

\newcommand{\sysstatempc}[2]{\sysstate{#1|#2}}
\newcommand{\sysinputmpc}[2]{\sysinput{#1|#2}}

\newcommand{\lengths}{\Lambda}

\newcommand{\mpccontr}[2]{\pi\ulm{MPC}(#1, #2)}

\newcommand{\aampccontr}[2]{\pi\ulm{AMPC}(#1, #2)}

\newcommand{\boreward}[1]{R(#1)}

\newcommand{\stagecost}[3]{\ell_{#1}(\kappa, \sysstatempc{#2}{#3}, \sysinputmpc{#2}{#3})}

\DeclareMathOperator*{\argmax}{arg\,max}

\newcommand{\ie}{i\/.\/e\/.,\/~}
\newcommand{\eg}{e\/.\/g\/.,\/~}
\newcommand{\cf}{cf\/.\/~}

\newcommand{\videolink}{at \href{https://youtu.be/EhMNIMqVKZk}{\color{blue}{\texttt{https://youtu.be/EhMNIMqVKZk}}}}
\newcommand{\codelink}{\url{https://github.com/hshose/BO-parameter-adaptive-AMPC}}

\newcommand{\fakepar}[1]{\vspace{0mm}\noindent\textit{#1.}}

\newcommand{\plotfontsize}{9pt}

\begin{document}
\title{Fine-Tuning of Neural Network Approximate MPC without Retraining via Bayesian Optimization
}
\titlerunning{Fine-Tuning Approximate MPC via BO}

\author{Henrik~Hose\inst{1}\and
Paul~Brunzema\inst{1} \and
Alexander~von~Rohr\inst{2} \and
Alexander~Gr{\"a}fe\inst{1} \and
Angela~P.~Schoellig\inst{2} \and
Sebastian~Trimpe\inst{1}}
\authorrunning{H. Hose et al.}
\institute{Institute for Data Science in Mechanical Engineering (DSME)\\
RWTH Aachen University, Germany\\
\email{henrik.hose@dsme.rwth-aachen.de} \and
Learning Systems and Robotics Lab\\Technical University of Munich, Germany%
}
\maketitle
\begin{abstract}
Approximate model-predictive control (AMPC) aims to imitate an MPC's behavior with a neural network, removing the need to solve an expensive optimization problem at runtime.
However, during deployment, the parameters of the underlying MPC must usually be fine-tuned.
This often renders AMPC impractical as it requires repeatedly generating a new dataset and retraining the neural network.
Recent work addresses this problem by adapting AMPC without retraining using approximated sensitivities of the MPC's optimization problem.
Currently, this adaption must be done by hand, which is labor-intensive and can be unintuitive for high-dimensional systems.
To solve this issue, we propose using Bayesian optimization to tune the parameters of AMPC policies based on experimental data.
By combining model-based control with direct and local learning, our approach achieves superior performance to nominal AMPC on hardware, with minimal experimentation.
This allows automatic and data-efficient adaptation of AMPC to new system instances and fine-tuning to cost functions that are difficult to directly implement in MPC.
We demonstrate the proposed method in hardware experiments for the swing-up maneuver on an inverted cartpole and yaw control of an under-actuated balancing unicycle robot, a challenging control problem.

\end{abstract}

\section{Introduction}
\label{sec:introduction}
Model-predictive control (MPC) is a modern optimization-based control method for nonlinear systems that provides theoretical guarantees for constraint satisfaction and stability~\citep{rawlings2017model}.
MPC has achieved remarkable results in practical robotics applications~\citep{ostafew2014learning,liniger2015optimization,williams2016aggressive, bledt2020extracting, song2023reaching}.
However, MPC requires solving an optimization problem periodically at runtime, making real-world deployment unfeasible for fast dynamical systems, even when dealing with mildly complicated dynamics, cost functions, and constraints.
Approximate MPC (AMPC) is one way to solve this challenge:
A fast-to-evaluate function approximator, typically a neural network (NN), is trained in an imitation-learning fashion on a large dataset of samples from the MPC, i.e., a dataset of states and optimal actions (see \citep{gonzalez2023neural} for a recent survey on AMPC with NNs).
Computing this large dataset can be done offline and in parallel on large computation clusters, but it can easily take tens of thousands of core-hours, especially for high-dimensional systems.
This poses a problem when deploying AMPC in practice, as often multiple iterations over parameter values of the MPC in the model, cost functions, and constraint sets are required to achieve the desired real-world control performance.
For every parameter change, the entire dataset must be regenerated.
In our opinion, this is one of the reasons why applications of AMPC in fast-moving dynamical and robotics systems are rare.
In a recent paper~\citep{hose2024parameter}, it is shown that a second neural network approximating the gradients of the optimal actions with respect to parameters of the MPC problem (also known as sensitivities of the MPC optimization problem) can be used to adapt an AMPC to changes in system parameters online -- without recomputing large datasets or training neural networks.
However, the current approach requires the parameters to be chosen by hand, making deployment of the AMPC labour intensive.

\begin{figure*}[t]
    \centering
    \resizebox{1.1\textwidth}{!}{
    \input{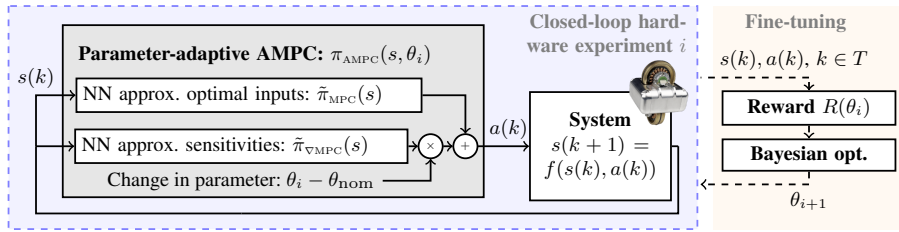}}
    \caption{Automatic tuning of parameter-adaptive AMPC with Bayesian optimization (BO). Approximate nominal MPC inputs are linearly adapted by approximate sensitivities to deviations from nominal parameters $\paramnom$. Parameters are directly tuned with a few experiments using BO, maximizing a closed-loop reward~$R$ on the real system.}
    \label{fig:method}
\end{figure*}
 
\fakepar{Contribution} 
We use Bayesian optimization (BO) to find the optimal parameterization for parameter-adaptive AMPCs based on closed-loop experiments~(Fig.~\ref{fig:method}) in a data-efficient manner.
The reward given to the BO reflects our true control objective and allows us to optimize the AMPC in this direction.
The true objective can be sparse in states and time, or binary such as success or failure, which is difficult to implement in MPC.
We show the effectiveness of our method in two hardware experiments: a common cartpole system and yaw control of a balancing reaction wheel unicycle robot (Fig.~\ref{fig:systems}), \cite{hose2025miniwheelbot}.
The latter problem is linearly uncontrollable, making it a challenging use case for nonlinear AMPC.
On both systems, we achieve stabilization and disturbance rejection in the closed loop after only a handful of experiments.
Automatic tuning is able to overcome model mismatch between simulation and hardware with only 20 experiments to achieve desirable performance.
In summary, our contributions are:
\begin{compactenum}
\item Automatic tuning of a parameter-adaptive AMPC to new system instances in a direct, data-driven manner using BO, and without retraining neural networks.
\item Fine-tuning of the parameter-adaptive AMPC to new reward functions that are difficult to implement with classic MPC (\eg due to sparsity).
\item Experimental validation on two unstable systems: a classic cartpole swing-up and stabilization task and yaw control of an underactuated reaction wheel unicycle robot. 
\end{compactenum}
A video of our experiments is available \videolink.

\section{Related Work}
\label{sec:related_work}
The method developed herein draws on two active research areas in robot learning: approximating MPC (\ie imitation learning from MPC) and BO for controller tuning. 
We review related works in each of these, highlighting how we combine the two in a new way.

\fakepar{Approximate MPC}
AMPC is a technique that finds a fast-to-evaluate but approximate explicit representation of a MPC through NN training~\citep{gonzalez2023neural}.
Unlike other explicit representations, for example, memory-intensive look-up tables~\citep{bayer2016tube,florence2022implicit} or explicit MPC for linear systems~\citep{bemporad2002explicit,alessio2009survey}, AMPC is applicable to general, nonlinear systems and requires only small NNs~\cite{nubert2020safe,carius2020mpc}.
These networks can be evaluated on small microcontrollers in milliseconds~\citep{hose2024parameter}.
Nonetheless, very few publications~\cite{nubert2020safe,carius2020mpc} apply AMPC in real-world robotics tasks, which we attribute to a significant practical issue:
Even though the dataset synthesis is performed offline, it can take tens of thousands of CPU core hours.
Additional computation time is required for training the NN approximation.
In classic AMPC, a new dataset must be computed and a new NN trained for every tuning iteration.
This is not practical in many applications as system instances have slightly different physical parameters, for example, masses, lengths, or friction parameters~\citep{adhau2019embedded}.

The first way to overcome this issue is to approximate a MPC that is robust against parameter uncertainties~\cite{nubert2020safe}, leading, however, to conservatism and requiring a-priori known uncertainty bounds. 
Alternatively, a nominal AMPC can be used to warm-start an optimizer online in hopes to speed up computations~\cite{klauvco2019machine,chen2022large};
this, however, is not always faster~\cite{vaupel2020accelerating} and much slower than NN inference.

An alternative is the recently introduced parameter-adaptive AMPC~\cite{hose2024parameter}, described in detail in Sec.~\ref{sec:method}.
It allows adapting the output of an AMPC to changes in MPC parameters (\eg parameters of the system dynamics model or cost function) with a locally linear predictor based on approximated sensitivities of the MPC's optimization problem.
Practical experiments indicate that this provides intuitive tuning nobs that generalize an AMPC to system instances with quite different parameters.
Further, the NNs required in parameter-adaptive AMPC are small enough to be evaluated in milliseconds on common microcontrollers that cost only a few dollars, making this method particularly appealing for real-world applications at scale.
However, the proposed method~\citep{hose2024parameter} relies on expert knowledge to tune the parameters correctly to achieve desired performance.
This manual approach can be cumbersome for systems with many parameters or mass production.
In this work, we showcase the efficiency of BO in automatically tuning parameter-adaptive AMPC for systems with many parameters through only a few hardware experiments.
We demonstrate this on an eleven-dimensional tuning task for yaw control of a unicycle robot, for which manual tuning as in~\cite{hose2024parameter} is infeasible.

\fakepar{Bayesian Optimization for MPC Tuning}
BO is a sample-efficient black-box optimization method~\cite{garnett2023bayesian} that gained popularity for automatic controller tuning in recent years~\cite{paulson2023tutorial}, for example, to tune the cost matrices in LQR control~\cite{marco2016automatic}, optimize gaits~\cite{calandra2016bayesian}, or the controller of a quadcopter \cite{berkenkamp2016safe}.
In the context of classic MPC, BO has been used to optimize hyperparameters~\cite{andersson2016model,guzman2020heteroscedastic,gharib2021multi,guzman2022bayesian}.
BO can also be used to tune the prediction model to optimize closed-loop performance~\cite{piga2019performance,sorourifar2021data}.
Crucially, all aforementioned BO methods rely on solving MPC optimization problems online at control frequency.
We overcome this problem by leveraging parameter-adaptive AMPC, which allows us to quickly obtain approximated optimal solutions through forward passes of the NNs even on low-cost hardware.

Local BO methods such as GIBO~\cite{muller2021local,wu2024behavior} and TuRBO~\cite{eriksson2019scalable} can cope with the increasing complexity of the optimization problem in higher dimensions.
Such approaches have proven to be especially useful in finding good optima in a data-efficient manner by restricting exploration to a local region.
As our approach focuses on fine-tuning the AMPC to the task at hand, we will resort to a local BO method, specifically TuRBO~\cite{eriksson2019scalable}, to find an optimal configuration of parameters used in the parameter-adaptive AMPC.

\section{Fine-Tuning of Parameter-Adaptive AMPC with Bayesian Optimization}
\label{sec:method}
In this section, we first describe how to define a set of parameterized policies using parameter-adaptive AMPC and, second, how to solve the policy search problem data-efficiently with BO.

We consider general, nonlinear, discrete-time dynamical systems
\begin{equation}\label{eqn:nominal_system}
    s(k+1) = f(s(k), a(k)), ~~\sysstate{0} = s_0,
\end{equation}
where $s$ are the states and $a$ the actions.
While we do not explicitly account for process and sensor noise in~\eqref{eqn:nominal_system}, the later hardware experiments naturally include such uncertainties.
The system~\eqref{eqn:nominal_system} is controlled by an AMPC policy $\pi_{\theta}$ parameterized with~$\theta$, \ie
$a(k) = \pi_{\theta}(s(k))$.
Here,~$\theta$ are parameters of the MPC that is imitated and explained in detail in~Sec.~\ref{ssec:param_mpc}.
The closed-loop system generates trajectories $\{(s(k),a(k))\}_{k=0}^{T}$ of length $T$ that depend on the policy parameters $\theta$.
The novelty and goal of this paper is to automatically fine-tune the AMPC policy $\pi_\theta$ such that
\begin{equation}\label{eq:policy_search}
    \pi_{\theta^*} = \argmax_{\pi_\theta \in \Pi_\mathrm{AMPC}} R(\theta),
\end{equation}
based on trajectories of the closed-loop system from hardware experiments.
We do not assume any properties of the reward $R$, for example, it can be sparse or non-Markovian.

We structure the rest of this section as follows:
Sec.~\ref{ssec:param_mpc} describes the parameterized MPC problem, which~$\pi_\theta$ imitates.
Then, Sec.~\ref{ssec:paampc} defines the search space $\Pi\ulm{\mathrm{AMPC}}$ in problem \eqref{eq:policy_search} as a parameter-adaptive AMPC~\cite{hose2024parameter} that keeps the parameterization in~$\theta$ intact.
Finally, in Sec.~\ref{ssec:bo}, we fine-tune~$\theta$ using local BO to find the optimal parameters for the AMPC, such that~$\pi_{\theta^*}$ is a solution to~\eqref{eq:policy_search}.

\subsection{Parameterized Model-Predictive Control}\label{ssec:param_mpc}
We formulate the following nonlinear MPC problem depending on parameters $\theta\in\Theta$
\begin{align}
	\begin{split}\label{eqn:implicit-mpc}
	&a_{\param}^* = \arg \min_{a}{\textstyle\sum_{\kappa=0}^{N}}\stagecost{{\param}}{\kappa}{k}\\
	\mathrm{s.t.~~} & \sysstatempc{0}{k} = \sysstate{k}, \quad \sysstatempc{\kappa+1}{k} = \Tilde{f}_{\param}(\sysstatempc{\kappa}{k}, \sysinputmpc{\kappa}{k}),\\
	& \sysstatempc{\kappa}{k} \in \mathcal{X}_{\param}(\kappa), \quad \sysinputmpc{\kappa}{k} \in \mathcal{U}_{\param}(\kappa) \quad \forall\kappa=0\dots N,
	\end{split}
\end{align}
where $\ell_{\param}$ is a cost function, and $\mathcal{X}_{\param}(\kappa)$ and $\mathcal{U}_{\param}(\kappa)$ are the state and input constraints. 
The loss function, constraint sets, and MPC's dynamics model $\Tilde{f}_{\param}$ may depend on the parameter vector $\param$ (\eg friction coefficients).

In classic MPC, \eqref{eqn:implicit-mpc} is solved repeatedly at every time~$k$ and the first element of the optimal predicted action sequence is applied to the system.
Thus, the optimization problem~\eqref{eqn:implicit-mpc} implicitly defines a mapping from states to actions, which we call the policy~$\mpccontr{\sysstate{k}}{\param}=a_{\param}^*(0|k)$.
The gradient of this policy with respect to the parameters at a specific state, $\tfrac{\partial}{\partial \param}\mpccontr{s(k)}{\param}|_{\theta\ulm{nom}}$, where $\paramnom$ are the nominal parameters, also known as sensitives, can be computed by commonly used NLP solvers along with~$a^*$~\cite{fiacco1976sensitivity, hart2017pyomo, andersson2016model, pirnay2012optimal}.

\subsection{Parameter-Adaptive AMPC}\label{ssec:paampc}

This section summarizes the AMPC control strategy with sensitivities from \citep{hose2024parameter}.
It makes it possible to locally adjust a neural network approximation of~\eqref{eqn:implicit-mpc} to parameters around nominal parameters.
To this end, parameter-adaptive AMPC combines two NNs to a single policy.
First, a neural network is trained to imitate the nominal policy $\pi\ulm{MPC}$ using a large dataset $\mathcal{D}\ulm{nom} =\{(s_j, \pi\ulm{MPC}(s_j,\paramnom))\}$. This yields the approximate nominal policy $\Tilde{\pi}\ulm{MPC}$.
Second, when computing the dataset $\mathcal{D}_{\mathrm{nom}}$ we also compute the sensitives which are collected in the dataset $\mathcal{D}\ulm{\nabla MPC} = \{(s_j, \tfrac{\partial}{\partial \param}\mpccontr{s_j}{\param}|_{\theta_{\scaleto{ \mathrm{nom} }{2pt}}}) \}$.
We train a neural network to approximate the sensitives as~$\Tilde{\pi}\ulm{\nabla AMPC}$.
The approximate sensitivities can be used as linear predictor around $\paramnom$ to adapt the optimal action given a change in the parameters $\theta$.
Thus, the parameter-adaptive AMPC policy is
\begin{equation}
    \aampccontr{s}{\param} = \Tilde{\pi}\ulm{MPC}(s) + \Tilde{\pi}\ulm{{\nabla \mathrm{MPC}}}(s) (\param - \paramnom).
\end{equation}
We define the policy search problem in \eqref{eq:policy_search} over the set of policies induced by parameter-adaptive AMPC and indexed by $\theta$ as
$ \Pi\ulm{\mathrm{AMPC}} = \{ \pi_{\theta} : s \mapsto \aampccontr{s}{\param} \}$.
In the next section, we use BO to automatically find the optimal parameters for a given reward function.

\subsection{Task-Specific Fine-Tuning with Bayesian Optimization}\label{ssec:bo}
We formalize the tuning problem \eqref{eq:policy_search} within the 
policy set $\Pi\ulm{\mathrm{AMPC}}$ as a black-box optimization problem
\begin{equation}\label{eq:objective}
    \param^* = \argmax_{\param \in \Theta} R( \param).
\end{equation}
Thus, the search over policies in~\eqref{eq:policy_search} reduces to finding optimal parameters~$\param^*$.
Importantly, the reward function $R$ and the MPC cost function~$\ell$ do not need to coincide.
We can tune an existing AMPC policy for new systems as well as fine-tune to specific tasks.
We use BO to solve the black-box optimization problem~\eqref{eq:objective}.
This will allow us to formulate a high-level reward function that might not be practical for classic MPC and then \emph{automatically} tune the sensitives to optimally solve the problem at hand.
In BO, a probabilistic model of the reward function~$R$, here a Gaussian process~(GP) as $\mathcal{GP}(m, k)$, where $m : \Theta \to \mathbb{R}$ and $k : \Theta \times \Theta \to \mathbb{R}$ are the mean and kernel function, respectively, and an acquisition function determine the next parameters~$\param_i$ to evaluate at iteration~$i$.
At each iteration, we conduct closed-loop experiments and collect noisy rewards as $R_i = R(\theta_i) + \epsilon_i$ with $\epsilon_i \sim \mathcal{N}(0, \sigma_n^2$) to sequentially build a data set $\mathcal{D}\ulm{opt} = \{(\theta_i, R_i)\}_{i=1}^{n}$ that is informative about the optimal policy. 
Here, $n$ denotes the number of experiments conducted thus far.
The noise with variance $\sigma_n^2$ might come from different initial states $s_i(0)$ or disturbances during the experiments.

We choose TuRBO~\cite{eriksson2019scalable} as our BO method to locally fine-tune the initial solution of the given task.
TuRBO maintains a trust region $\mathcal{TR}_i$ as the hyperbox around the parameter of the best observed value $\hat{\param}^*_i$ based on the current GP lengthscales $\lengths_i \in \mathbb{R}_+^d$ and the current base length $L_{\ulm{TR}}$ as 
\begin{align}
    \mathcal{TR}_i = \left\{ \param \in \Theta \Bigm| \vert \param_j - \hat{\param}^*_{i,j} \vert \leq \frac{L_j}{2}, \, \forall j \in [d] \right\}  \, \text{where}\, L_j = L_{\ulm{TR}} \frac{\lambda_j}{\left(\prod_{p=1}^d \lambda_p \right)^{1/d}}.
\end{align}
Here, $\lambda_j$ and $\param_j$ are the $j$-th entry in $\Lambda_i$ and $\param$, respectively.
With this formulation, the trust region scales anisotropically with respect to the GP kernel lengthscales while ensuring that the hypervolume at each iteration is at most $L_{\ulm{TR}}^d$.
As acquisition function, TuRBO uses Thompson sampling.
At each iteration, we generate a realization of the posterior GP that is conditioned on $\mathcal{D}\ulm{opt}$ and trained through maximum likelihood estimation, and choose as the parameters for the next iteration as
\begin{align}
    \param_{i+1} = \arg\max_{\param \in \mathcal{TR}_i} &\hat{R} \text{ where } \hat{R} \sim \mathcal{GP}_{\mathcal{D}\ulm{opt}}(\mu(\param), \sigma^2 (\param)) \\
    \text{and }\,\, \mu(\param) &= k(\param, X) K^{-1} y, \\
    \sigma^2(\param) &= k(\param, \param) - k(\param, X) K^{-1} k(X, \param).
\end{align}
Here, $X = [\param_1, \dots, \param_n]$ is the matrix of observed points, $y = [R_1, \dots, R_n]$ is the corresponding vector of reward realizations, and $K = k(X, X) + \sigma_n^2 I$ is the Gram matrix.
Since a posterior sample of a GP can not be optimized numerically, we follow \cite{eriksson2019scalable} and generate a candidate set of solutions using a Sobol sequence within $\mathcal{TR}_i$ and evaluate the posterior sample on the finite candidates.
We then simply choose the candidate with the highest predicted reward.
The trust region is updated dynamically similar to \cite{nelder1965simplex}: If $\param_{i+1}$ repeatedly improves the best value, increase $L_{\ulm{TR}}$ as $L_{\ulm{TR}} \leftarrow \min(2 L_{\ulm{TR}}, L_{\ulm{max}})$, if no improvement is achieved for some iterations, reduce $L_{\ulm{TR}}$ as $L_{\ulm{TR}} \leftarrow L_{\ulm{TR}}/ 2$, and if $L_{\ulm{TR}} < L_{\ulm{min}}$, reset the algorithm.
We fully utilize the local idea of TuRBO by only considering one trust region that shrinks over time and collapses to the locally optimal solution.
This local approach is crucial because it does not require explicit parameter bounds~$\Theta$; the bound of the local trust region is inferred based on the length scales of the kernel of the GP.
Still, bounds can help to further reduce the search space and possibly the volume of the hyper box $\mathcal{TR}$~(\cf \cite[Sec.~2]{eriksson2019scalable}).

\section{Implementation on Benchmark Systems}
\label{sec:systems}
\begin{figure*}[t!]
    \centering
    \begin{subfigure}[b]{0.27\textwidth}
        \centering
        \includegraphics[width=0.945\linewidth]{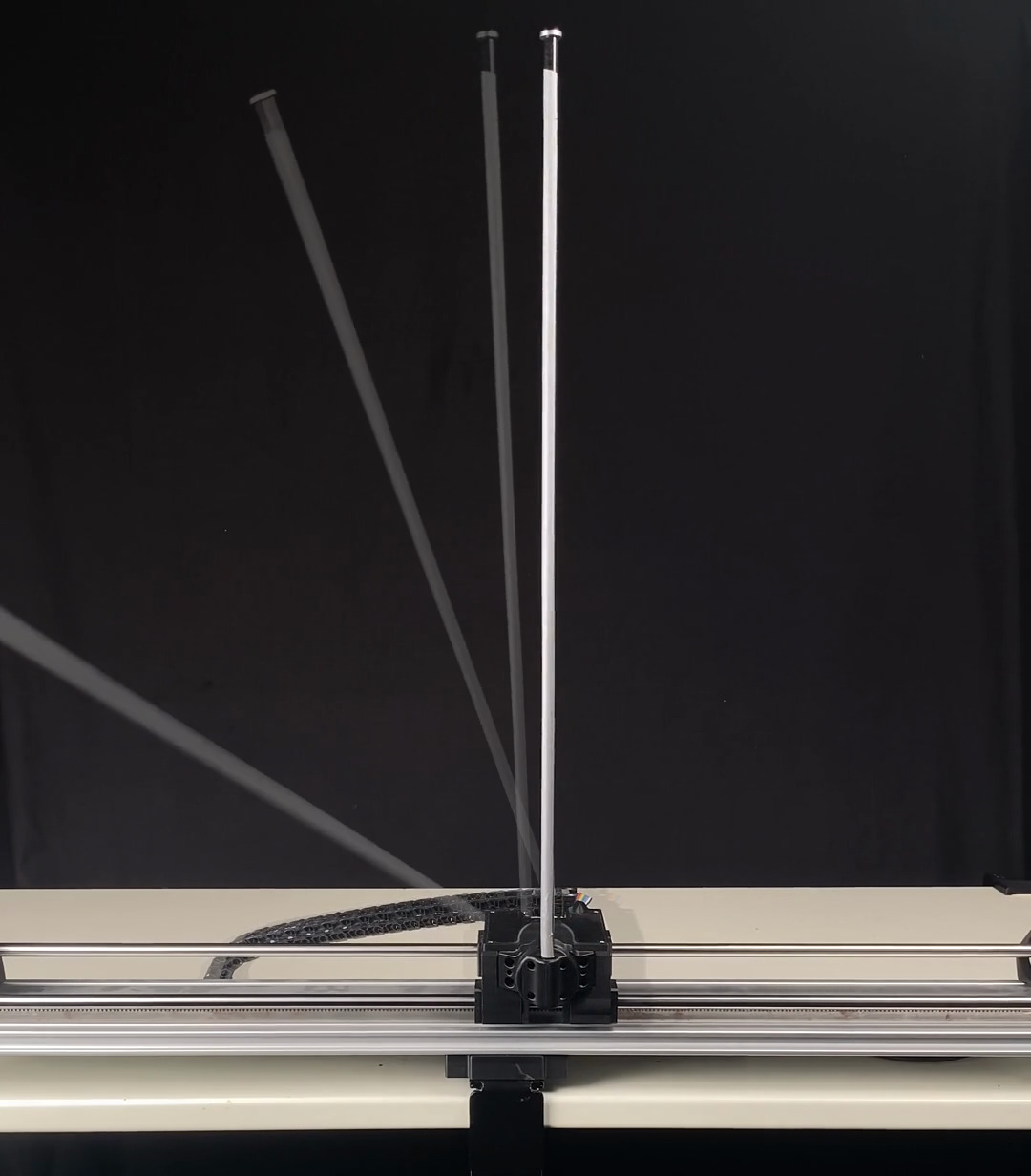}
        \caption{Cartpole}
    \end{subfigure}
    \begin{subfigure}[b]{0.35\textwidth}
    \centering
        \input{figures/systems/rotation_wb.tex}%
        \caption{Mini Wheelbot (side view)}
    \end{subfigure}
    \begin{subfigure}[b]{0.35\textwidth}
    \centering
        \input{figures/systems/mosaic_wb.tex}%
        \caption{Mini Wheelbot (top view)}
    \end{subfigure}
    \caption{Hardware systems used for evaluation: cartpole and reaction wheel unicycle robot.
    The cartpole is a classic control benchmark system on which we perform swing-up and stabilization. The Mini Wheelbot is a reaction wheel unicycle robot on which we demonstrate balancing and yaw control.
    A video of our experiments is available \videolink.}
    \label{fig:systems}
\end{figure*}

We implement and evaluate parameter-adaptive AMPC and tuning with BO on two benchmark systems: a cartpole and a reaction wheel unicycle robot~(Fig.~\ref{fig:systems}).
We chose both systems because they exhibit strong nonlinear dynamics, are unstable, and require fast feedback control, making them predestined for AMPC.
Further, we implement the parameter-adaptive AMPC (\ie neural network controller inference) on the onboard embedded CPUs\footnote{The cartpole has a STM32G474 ARM Cortex-M4 CPU running at~\SI{170}{\mega\hertz}. The Mini Wheelbot has a Raspberry Pi CM4 with BCM2711 quad-core Cortex-A72 CPU running at~\SI{1.5}{\giga\hertz}}.
The CPUs are not powerful enough to solve the implicit MPC optimization problem in real-time.
While the NN controllers run on embedded CPUs, we conveniently use BoTorch~\cite{balandat2020botorch} on a laptop.
Our implementation is publicly available\footnote{\codelink}.

\subsection{Cartpole Swing-Up and Stabilization}
The cartpole system is a standard benchmark in control~\citep{boubaker2013inverted}.
We use a single policy to control the swing-up and stabilization of the pole in the upwards-facing position without violating the constraints on the rail.
The AMPC implementation closely resembles the publicly available code used  in~\cite{hose2024parameter}.
Therefore, we only elaborate on the fine-tuning. 

The cartpole's state consists of cart position, pendulum angle, and their derivatives, thus~$s\ulm{cartpole}\in\mathbb{R}^4$ with voltage applied to the cart's motor as action~$a\ulm{cartpole}\in\mathbb{R}$.
The dynamics function used in the MPC optimization problem is parameterized by~$\theta\ulm{cartpole}=[m_\text{add},M,C_1, C_2, C_3]\in\mathbb{R}^5$, where $m_\text{add}$ is the mass atop the rod, $M$ is the mass of the cart, and $C$ are friction and motor constants.

\fakepar{Reward for Fine-Tuning}
The task of the pendulum is to perform a swing-up as fast as possible and stabilize afterwards around the upright position with the cart at the center of the rail for a total of \SI{20}{\second}.
We formalize this in the following sparse reward function:
$
    R\ulm{cartpole}(\param) = \frac{1}{T} t\ulm{up} - \frac{w\ulm{pos}}{T-t\ulm{up}}\sum_{k=t\ulm{up}}^T  s\ulm{pos}(k)^2,
$
where $t\ulm{up}$ is the number of time steps that the pendulum has successfully remained in the upright position, \ie the angle remains within $[-\SI{15}{\degree}, \SI{15}{\degree}]$.
If no swing-up is achieved or constraints are violated ($|s\ulm{pos}|>\bar{s}\ulm{pos}$), the reward is set to $0$.
All weighting factors are in Appendix~\ref{app:hyperparameters}.
Implementing such an objective would be difficult in classic MPC.

\subsection{Yaw Control on a Balancing Reaction Wheel Robot}
The Mini Wheelbot is a symmetric, balancing, reaction wheel unicycle robot~\citep{ hose2025miniwheelbot} with two wheels:
one driving wheel and an orthogonal reaction wheel.
The robot can directly control its pitch and roll angles by applying torques to its wheels.
However, the robot does not have a third ``turntable'' actuator to control its yaw directly.
Classic linear control methods fail at controlling the yaw angle for this class of robots~\citep{lee2012decoupled, neves2021discrete,rosyidi2016speed}.
However, a nonlinear MPC, as in this paper, can use the reaction wheel's nonlinear gyroscopic effects to steer the robot's orientation.

\fakepar{MPC Implementation} The robot's state can be described by minimal coordinates consisting of roll, pitch, and yaw orientation, and both wheel encoder values, and all of their derivatives~\citep{geist2022wheelbot}, therefore~$s\ulm{wheelbot}\in\mathbb{R}^{10}$.
The actions are the torques applied by the motors to both wheels,~$a\ulm{wheelbot}
~\in~\mathbb{R}^2$.
The robot's continuous-time, nonlinear dynamics in implicit form are described in detail in~\citep{daud2017dynamic}.
The dynamics are parametrized by~$\theta\ulm{wheelbot}=[m_\text{B},m_\text{W,R},I_\text{B},I_\text{W,R}, r_\text{W,R}, l_\text{WB}, \mu]$, where~$m$ denotes masses, $I\in\mathbb{R}^3$ diagonals of mass moment of inertia matrices,~$r$ the effective wheel radius,~$l$ the distance between the wheels' rotation axis, and~$\mu\in\mathbb{R}^2$ friction parameters of the wheel-to-ground contact, and indices~$\text{W}$ and $\text{R}$ the driving and reaction wheels, and~$\text{B}$ the robots main body.
Due to symmetries, the number of free parameters is~$\theta\ulm{wheelbot}\in\mathbb{R}^{11}$.
We implement a nonlinear MPC with quadratic cost, a horizon lookahead of~\SI{1.2}{\second} discretized with~\SI{20}{\milli\second} steps using the implicit integrators from~\cite{frey2023fast}, and appropriate action, state, and terminal constraints.
The MPC optimization problem is formulated in CasADi~\cite{andersson2019casadi} with sensitivities by~\cite{andersson2018sensitivity}, and solved with IPOPT~\cite{wachter2006implementation}.

\fakepar{Neural Network Approximation}
The dataset that we synthesize contains~\num{3.5} million random samples of states and optimal actions.
Computation of the dataset takes \num{86} thousand core hours\footnote{computed in parallel on Intel Xeon 8468 CPUs at \SI{3.8}{\giga\hertz}}.
We use fully connected feedforward NNs with 100 neurons per layer, a mixture of tangent hyperbolic and rectified linear activations, and 4 layers and 8 layers for approximating inputs and sensitivities, respectively.
We implement the NN inference in C++ with Eigen on the Mini Wheelbot's onboard CPU\footnotemark[1].
Inference on both NNs takes less than~\SI{300}{\micro\second}; thus, we can evaluate the AMPC at a control frequency of~\SI{200}{\hertz}.

\fakepar{Reward for Fine-Tuning} 
The objective of the Mini Wheelbot is to control its yaw to a sequence of~\num{4} setpoints with~\SI{90}{\degree} step responses while balancing in place.
Every episode takes \SI{22}{\second}.
We choose a sparse reward for fine-tuning that only considers the error in yaw, roll, and pitch angles~$s\ulm{yrp}$ and driving wheel angle~$s\ulm{wheel}$ as
\begin{equation}
    R\ulm{wheelbot}(\param)~=~-\tfrac{1}{T}~\textstyle\sum_{k=0}^T~w\ulm{yrp}^\top(s\ulm{yrp}(k)-s\ulm{yrp,ref}(k))^2 + w\ulm{wheel} s\ulm{wheel}(k)^2,
\end{equation}
where~$w$ are appropriate weights, to let the robot reorient in place effectively.
A failed experiment in which the robot crashes yields a reward of $-1$.
This is two times smaller than the worst reward achieved without a crash.

\section{Experimental Results}
\label{sec:results}
In this section, we present the results from simulation and hardware experiments on automatically tuning the parameter-adaptive AMPC with~BO for the two systems presented in Sec.~\ref{sec:systems}.
The simulations' primary goal is to evaluate our method's generalizability.
In the hardware experiments, we aim to demonstrate that our method is capable of running in real-world conditions, on low-cost hardware, and can learn efficiently within a feasible amount of time.

\begin{figure}[t]
    \fontsize{\plotfontsize}{\plotfontsize}\selectfont
    \centering
    \newcommand{\plotheight}{1.3in}
    \newcommand{\plotwidth}{2.5in}
    \begin{subfigure}[b]{0.49\linewidth}
      \centering
      \begin{tikzpicture}
  \definecolor{darkgray100}{RGB}{100,100,100}
  \definecolor{darkgray176}{RGB}{176,176,176}
  \definecolor{lightgray204}{RGB}{204,204,204}
  
  \begin{axis}[
      name=reward,
      xshift=0mm,
      anchor=north,
      xmax=18, 
      xmin=0,
      ymin=-0.1,ymax=0.9, 
      y grid style={darkgray176},
      y label style={yshift=-6mm},
      x label style={yshift=2mm},
      ytick={0, 0.2, 0.4, 0.6, 0.8},
      yticklabels={0, \phantom{0}, \phantom{0}, \phantom{0}, 0.8},
      label style={font=\footnotesize},
      ticklabel style={font=\scriptsize},
      ylabel={{Reward~$R$}},
      xlabel={Episode},
      height=\plotheight, width=\plotwidth,
      legend cell align={left},
      legend style={
        fill opacity=1,
        draw opacity=1,
        text opacity=1,
        at={(1.1,-0.18)},
        anchor=south east,
        draw=lightgray204,
        font=\scriptsize
      },
      reverse legend
      ]

      \addplot[name path=bo_top,color=brown!70, forget plot] table [x=epoch, y=max, col sep=comma] {figures/simulationresults/cartpole_simulation_random_mean_min_max_reward.csv};
      \addplot[name path=bo_down,color=brown!70, forget plot] table [x=epoch, y=min, col sep=comma] {figures/simulationresults/cartpole_simulation_random_mean_min_max_reward.csv};
      \addplot[brown!50,fill opacity=0.5, forget plot] fill between[of=bo_top and bo_down];
      \addplot[name path=bo_down, very thick, color=brown] table [x=epoch, y=mean, col sep=comma] {figures/simulationresults/cartpole_simulation_random_mean_min_max_reward.csv};
      \addlegendentry{Sobol sampling};
      
      \addplot[mark=none, very thick, color=black!50, dashed] table [x=epoch, y=mean_initial, col sep=comma] {figures/simulationresults/cartpole_simulation_turbo_mean_min_max_reward.csv};
      \addlegendentry{initial parameters};  
  
      \addplot[name path=bo_top,color=blue!70, forget plot] table [x=epoch, y=max, col sep=comma] {figures/simulationresults/cartpole_simulation_turbo_mean_min_max_reward.csv};
      \addplot[name path=bo_down,color=blue!70, forget plot] table [x=epoch, y=min, col sep=comma] {figures/simulationresults/cartpole_simulation_turbo_mean_min_max_reward.csv};
      \addplot[blue!50,fill opacity=0.5, forget plot] fill between[of=bo_top and bo_down];
      \addplot[name path=bo_down, very thick, color=blue] table [x=epoch, y=mean, col sep=comma] {figures/simulationresults/cartpole_simulation_turbo_mean_min_max_reward.csv};
      \addlegendentry{\textbf{Bayesian optim.}};  
  \end{axis}
\end{tikzpicture}%
      \caption{Cartpole}
    \end{subfigure}
    \begin{subfigure}[b]{0.49\linewidth}
      \centering
      \begin{tikzpicture}
    \definecolor{darkgray100}{RGB}{100,100,100}
    \definecolor{darkgray176}{RGB}{176,176,176}
    \definecolor{lightgray204}{RGB}{204,204,204}

    \begin{axis}[
        name=reward,
        xshift=0mm,
        anchor=north,
        xmax=18, 
        xmin=0,
        ytick distance = 0.05,
        ymax=-0.07, ymin=-0.21,
        ytick={-0.2, -0.15, -0.1},
        yticklabels={-0.2, \phantom{0}, -0.1},
        y grid style={darkgray176},
        y label style={yshift=-6mm},
        x label style={yshift=2mm},
        label style={font=\footnotesize},
        ticklabel style={font=\scriptsize},
        xlabel={Episode},
        height=\plotheight, width=\plotwidth,
        legend cell align={left},
        legend style={
          fill opacity=1.0,
          draw opacity=1,
          text opacity=1,
          at={(1.1,-0.18)},
          anchor=south east,
          draw=lightgray204,
          font=\scriptsize
        },
        reverse legend
        ]
        
    \addplot[name path=bo_top,color=brown!70, forget plot] table [x=epoch, y=max, col sep=comma] {figures/hardwareresults/wheelbot_hardware_random_mean_min_max_reward.csv};
    \addplot[name path=bo_down,color=brown!70, forget plot] table [x=epoch, y=min, col sep=comma] {figures/hardwareresults/wheelbot_hardware_random_mean_min_max_reward.csv};
    \addplot[brown!50,fill opacity=0.5, forget plot] fill between[of=bo_top and bo_down];
    \addplot[name path=bo_down, very thick, color=brown] table [x=epoch, y=mean, col sep=comma] {figures/hardwareresults/wheelbot_hardware_random_mean_min_max_reward.csv};
    \addlegendentry{Sobol sampling};
    
    \addplot[mark=none, very thick, color=black!50, dashed] table [x=epoch, y=mean_initial, col sep=comma] {figures/hardwareresults/wheelbot_hardware_turbo_mean_min_max_reward.csv};
    \addlegendentry{initial parameters};  

    \addplot[name path=bo_top,color=blue!70, forget plot] table [x=epoch, y=max, col sep=comma] {figures/hardwareresults/wheelbot_hardware_turbo_mean_min_max_reward.csv};
    \addplot[name path=bo_down,color=blue!70, forget plot] table [x=epoch, y=min, col sep=comma] {figures/hardwareresults/wheelbot_hardware_turbo_mean_min_max_reward.csv};
    \addplot[blue!50,fill opacity=0.5, forget plot] fill between[of=bo_top and bo_down];
    \addplot[name path=bo_down, very thick, color=blue] table [x=epoch, y=mean, col sep=comma] {figures/hardwareresults/wheelbot_hardware_turbo_mean_min_max_reward.csv};
    \addlegendentry{\textbf{Bayesian optim.}};  
  \end{axis}
\end{tikzpicture}%
      \caption{Mini Wheelbot}
    \end{subfigure}
  \caption{Simulation results: Average, minimum, and maximum reward improvement on~\num{100} random, simulated systems stabilized by the same parameter-adaptive AMPC (no retraining of neural networks).
  Using Bayesian optimization (\textcolor{blue}{blue}) reliably improves performance with a sparse closed-loop objective given a rough initial guess. We include a pseudo-random baseline for comparison (\textcolor{brown}{brown}).}
  \label{fig:simulationresults}
\end{figure}
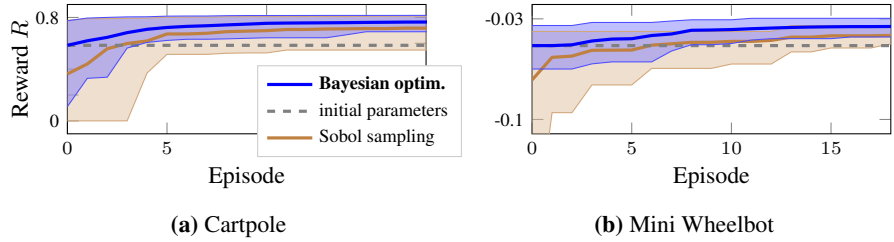

\subsection{Results in Simulation Experiments}
For the simulations, we generate \num{100} system instances with randomly sampled parameters for cartpole and Mini Wheelbot.
We then perform local fine-tuning starting from the nominal parameter values as initial conditions.
We use the same neural networks for all system instances, thus no retraining takes place.
In all experiments, we add quasi-random Sobol sampling within reasonable bounds around the nominal parameters as the baseline.
Sobol sampling is more sophisticated than grid search or random sampling and represents an engineering approach for finding good-performing parameters, providing a benchmark against which we can compare the sample efficiency of our method.

Fig.~\ref{fig:simulationresults} shows the results of our simulations for cartpole and Mini Wheelbot.
In both examples, the same neural network controller is able to stabilize a broad range of random systems, \ie the minimum and average reward indicates that the parameter-adaptive AMPC generalizes across instances of the same system with different parameters, such as mass or friction.
BO consistently improves over the initial parameters.
The Sobol sampling baseline also finds stabilizing parameters but requires more trials.

\begin{figure}[t]
    \fontsize{\plotfontsize}{\plotfontsize}\selectfont
    \centering
    \newcommand{\plotheight}{1.4in}
    \newcommand{\plotwidth}{2.5in}
    \begin{subfigure}[b]{0.49\linewidth}
      \centering
      \begin{tikzpicture}
    \definecolor{darkgray100}{RGB}{100,100,100}
    \definecolor{darkgray176}{RGB}{176,176,176}
    \definecolor{lightgray204}{RGB}{204,204,204}

    \begin{axis}[
        name=reward,
        xshift=0mm,
        anchor=north,
        xmax=18, 
        xmin=0,
        ymin=-0.1,ymax=0.9, 
        y grid style={darkgray176},
        y label style={yshift=-6mm},
        x label style={yshift=2mm},
        label style={font=\footnotesize},
        ticklabel style={font=\scriptsize},
        ytick={0, 0.2, 0.4, 0.6, 0.8},
        yticklabels={0, \phantom{0}, \phantom{0}, \phantom{0}, 0.8},
        xlabel={Episode},
        ylabel={{Reward $R$}},
        height=\plotheight, width=\plotwidth,
        legend cell align={left},
        legend style={
          fill opacity=0.8,
          draw opacity=1,
          text opacity=1,
          at={(0.95,0.05)},
          anchor=south east,
          draw=lightgray204,
          font=\scriptsize
        },
        reverse legend
        ]

        \addplot[name path=bo_top,color=brown!70, forget plot] table [x=epoch, y=max, col sep=comma] {figures/hardwareresults/cartpole_hardware_random_mean_min_max_reward.csv};
        \addplot[name path=bo_down,color=brown!70, forget plot] table [x=epoch, y=min, col sep=comma] {figures/hardwareresults/cartpole_hardware_random_mean_min_max_reward.csv};
        \addplot[brown!50,fill opacity=0.5, forget plot] fill between[of=bo_top and bo_down];
        \addplot[name path=bo_down, very thick, color=brown] table [x=epoch, y=mean, col sep=comma] {figures/hardwareresults/cartpole_hardware_random_mean_min_max_reward.csv};
        
        \addplot[mark=none, very thick, color=black!50, dashed] table [x=epoch, y=mean_initial, col sep=comma] {figures/hardwareresults/cartpole_hardware_turbo_mean_min_max_reward.csv};
    
        \addplot[name path=bo_top,color=blue!70, forget plot] table [x=epoch, y=max, col sep=comma] {figures/hardwareresults/cartpole_hardware_turbo_mean_min_max_reward.csv};
        \addplot[name path=bo_down,color=blue!70, forget plot] table [x=epoch, y=min, col sep=comma] {figures/hardwareresults/cartpole_hardware_turbo_mean_min_max_reward.csv};
        \addplot[blue!50,fill opacity=0.5, forget plot] fill between[of=bo_top and bo_down];
        \addplot[name path=bo_down, very thick, color=blue] table [x=epoch, y=mean, col sep=comma] {figures/hardwareresults/cartpole_hardware_turbo_mean_min_max_reward.csv};
    \end{axis}
\end{tikzpicture}%
      \caption{Cartpole}
    \end{subfigure}
    \begin{subfigure}[b]{0.49\linewidth}
      \centering
      \begin{tikzpicture}
    \definecolor{darkgray100}{RGB}{100,100,100}
    \definecolor{darkgray176}{RGB}{176,176,176}
    \definecolor{lightgray204}{RGB}{204,204,204}

    \begin{axis}[
        name=reward,
        xshift=0mm,
        anchor=north,
        xmax=18, 
        xmin=0,
        ytick distance = 0.05,
        ymax=-0.07, ymin=-0.21,
        ytick={-0.2, -0.15, -0.1},
        yticklabels={-0.2, \phantom{0}, -0.1},
        y grid style={darkgray176},
        y label style={yshift=-6mm},
        x label style={yshift=2mm},
        label style={font=\footnotesize},
        ticklabel style={font=\scriptsize},
        xlabel={Episode},
        height=\plotheight, width=\plotwidth,
        legend cell align={left},
        legend style={
          fill opacity=1.0,
          draw opacity=1,
          text opacity=1,
          at={(1.1,-0.18)},
          anchor=south east,
          draw=lightgray204,
          font=\scriptsize
        },
        reverse legend
        ]
        
    \addplot[name path=bo_top,color=brown!70, forget plot] table [x=epoch, y=max, col sep=comma] {figures/hardwareresults/wheelbot_hardware_random_mean_min_max_reward.csv};
    \addplot[name path=bo_down,color=brown!70, forget plot] table [x=epoch, y=min, col sep=comma] {figures/hardwareresults/wheelbot_hardware_random_mean_min_max_reward.csv};
    \addplot[brown!50,fill opacity=0.5, forget plot] fill between[of=bo_top and bo_down];
    \addplot[name path=bo_down, very thick, color=brown] table [x=epoch, y=mean, col sep=comma] {figures/hardwareresults/wheelbot_hardware_random_mean_min_max_reward.csv};
    \addlegendentry{Sobol sampling};
    
    \addplot[mark=none, very thick, color=black!50, dashed] table [x=epoch, y=mean_initial, col sep=comma] {figures/hardwareresults/wheelbot_hardware_turbo_mean_min_max_reward.csv};
    \addlegendentry{initial parameters};  

    \addplot[name path=bo_top,color=blue!70, forget plot] table [x=epoch, y=max, col sep=comma] {figures/hardwareresults/wheelbot_hardware_turbo_mean_min_max_reward.csv};
    \addplot[name path=bo_down,color=blue!70, forget plot] table [x=epoch, y=min, col sep=comma] {figures/hardwareresults/wheelbot_hardware_turbo_mean_min_max_reward.csv};
    \addplot[blue!50,fill opacity=0.5, forget plot] fill between[of=bo_top and bo_down];
    \addplot[name path=bo_down, very thick, color=blue] table [x=epoch, y=mean, col sep=comma] {figures/hardwareresults/wheelbot_hardware_turbo_mean_min_max_reward.csv};
    \addlegendentry{\textbf{Bayesian optim.}};  
  \end{axis}
\end{tikzpicture}%
      \caption{Mini Wheelbot}
    \end{subfigure}
    \caption{Hardware results: Average, min., and max. reward improvement in in hardware experiments.
    Bayesian optimization (\textcolor{blue}{blue}) tunes the AMPC to satisfactory performance in~\num{20} hardware experiments (``Episodes''). For comparison, we include a pseudo-random baseline (\textcolor{brown}{brown}).
    }
    \label{fig:hardwareresults:improvement}
\end{figure}
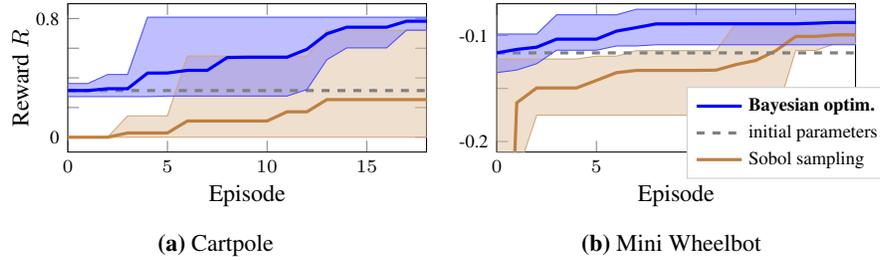

\subsection{Results in Hardware Experiments}

We deploy our method on the hardware systems shown in Fig.~\ref{fig:systems} to demonstrate that the proposed method can tune AMPC controllers without retraining, thus compensating for inevitable mismatch between nominal model (used for the MPC) and real hardware.
The actual system parameters for both systems differ from the nominal ones used in the MPC, which necessitates parameter-adaptive AMPC.
We initialize BO with parameters that successfully complete the task, \ie the cartpole can perform a swing-up and the Mini Wheelbot stabilizes and controls the yaw orientation.
In practical applications, the engineer would typically have good intuition about nominal parameters through direct measurements or average values from other system instances.
If such a good initial guess is not available, a set of random trials could also be used to initialize learning.

\begin{figure}[bt]
    \fontsize{\plotfontsize}{\plotfontsize}\selectfont
    \centering
    \newcommand{\plotheight}{1.7in}
    \newcommand{\plotheighthalf}{1.1in}
    \newcommand{\plotwidth}{2.5in}
    \begin{subfigure}[b]{0.48\linewidth}
        \centering
        \begin{tikzpicture}
\definecolor{darkgray100}{RGB}{100,100,100}
\definecolor{darkgray176}{RGB}{176,176,176}
\definecolor{lightgray204}{RGB}{204,204,204}

    \begin{axis}[
        name=pend,
        anchor=north,
        ytick style={color=black},
        ytick={-3.14, 0, 3.14},
        yticklabels={-$\pi$, $0$, $\pi$},
        ymajorgrids,
        y label style={yshift=-6mm},
        x grid style={darkgray176},
        xmajorgrids,
        xmin=0, xmax=20,
        xtick style={color=black},
        xticklabels={},
        y grid style={darkgray176},
        ylabel={{Ang [rad]}},
        height=\plotheighthalf, width=\plotwidth,
        label style={font=\scriptsize},
        ticklabel style={font=\scriptsize},
        legend cell align={left},
        legend style={
          fill opacity=1,
          draw opacity=1,
          text opacity=1,
          at={(1.05,0.2)},
          anchor=south east,
          legend columns=1,
          font=\scriptsize,
          /tikz/every even column/.append style={yshift=-2pt},
          /tikz/column 1/.append style={yshift=-2pt},
          inner sep=1pt,
          row sep=-2pt,
        },
        reverse legend,
    ]

    \addplot [mark=none, thick, color=darkgray100] coordinates {
        (0, 0) (22, 0)
    };    
    \addlegendentry{reference};

    \addplot[mark=none, thick, color=red] table [x=time, y=angle_pendulum, col sep=comma] {figures/hardwareresults/cartpole_hardware_turbo_initial.csv};
    \addlegendentry{initial};

    \addplot[mark=none, thick, color=blue] table [x=time, y=angle_pendulum, col sep=comma] {figures/hardwareresults/cartpole_hardware_turbo_best.csv};
    \addlegendentry{\textbf{optimized}};

    \end{axis}

    \begin{axis}[
        name=cart,
        at={(pend.below south)},
        xshift=0mm,
        anchor=north,
        ytick style={color=black},
        ytick={-0.4, 0, 0.4}, 
        yticklabels={-0.4, \small{$0$}, 0.4},
        ymajorgrids,
        ymin=-0.5, ymax=0.5,
        y label style={yshift=-6mm},
        x grid style={darkgray176},
        xmajorgrids,
        xmin=0, xmax=20,
        xtick style={color=black},
        y grid style={darkgray176},
        x label style={yshift=2mm},
        xlabel={Time [s]},
        ylabel={{Cart [m]}},
        label style={font=\scriptsize},
        ticklabel style={font=\scriptsize},
        height=\plotheighthalf, width=\plotwidth,
        ]

    \addplot[mark=none, thick, color=darkgray100] coordinates {
        (0, 0) (22, 0)
    };

    \addplot[mark=none, thick, color=red] table [x=time, y=position_cart, col sep=comma] {figures/hardwareresults/cartpole_hardware_turbo_initial.csv};

    \addplot[mark=none, thick, color=blue] table [x=time, y=position_cart, col sep=comma] {figures/hardwareresults/cartpole_hardware_turbo_best.csv};

\end{axis}
\node[above,font=\bfseries,xshift=3mm] at (current bounding box.north) {Cartpole};
\end{tikzpicture}%
      \vspace{-2em}
    \end{subfigure}
    \begin{subfigure}[b]{0.48\linewidth}
        \centering
        \begin{tikzpicture}[spy using outlines={circle, magnification=2, connect spies}]
\definecolor{darkgray100}{RGB}{100,100,100}
\definecolor{darkgray176}{RGB}{176,176,176}
\definecolor{lightgray204}{RGB}{204,204,204}

    \begin{axis}[
        name=yaw,
        anchor=north,
        ytick style={color=black},
        ytick={-1.57, 0, 1.57},
        yticklabels={-$\frac{\pi}{2}$, $0$, $\frac{\pi}{2}$},
        ymajorgrids,
        ymin=-2, ymax=2,
        y label style={yshift=-6mm},
        x grid style={darkgray176},
        xmajorgrids,
        xmin=0, xmax=22,
        xtick style={color=black},
        xticklabels={},
        y grid style={darkgray176},
        ylabel={{Yaw [rad]}},
        height=\plotheighthalf, width=\plotwidth,
        label style={font=\scriptsize},
        ticklabel style={font=\scriptsize},
        legend cell align={left},
        legend style={
          fill opacity=0.8,
          draw opacity=1,
          text opacity=1,
          at={(1.1,0.4)},
          anchor=south east,
          draw=lightgray204,
          font=\scriptsize
        },
        reverse legend,
    ]

    \addplot[mark=none, thick, color=darkgray100] table [x=time, y=yaw_setpoint, col sep=comma] {figures/hardwareresults/wheelbot_hardware_turbo_best.csv};

    \addplot[mark=none, thick, color=red] table [x=time, y=yaw, col sep=comma] {figures/hardwareresults/wheelbot_hardware_turbo_initial.csv};

    \addplot[mark=none, thick, color=blue] table [x=time, y=yaw, col sep=comma] {figures/hardwareresults/wheelbot_hardware_turbo_best.csv};

    \coordinate (spypoint1) at (axis cs:9.5,0); %
    \coordinate (magnifyglass1) at (axis cs:20,2.5);
    \spy [gray, very thick, size=1.2cm, every spy on node/.append style={thick}] on (spypoint1) in node[fill=white, ultra thick] at (magnifyglass1);
    
    \end{axis}

    \begin{axis}[
        name=drivewheel,
        at={(yaw.below south)},
        xshift=0mm,
        anchor=north,
        ytick style={color=black},
        ytick={-1.57, 0, 1.57},
        yticklabels={-$\frac{\pi}{2}$, \small{$0$}, $\frac{\pi}{2}$},
        ymajorgrids,
        ymin=-2, ymax=2,
        y label style={yshift=-6mm},
        x grid style={darkgray176},
        xmajorgrids,
        xmin=0, xmax=22,
        xtick style={color=black},
        y grid style={darkgray176},
        x label style={yshift=2mm},
        xlabel={Time [s]},
        ylabel={{Wheel [rad]}},
        label style={font=\scriptsize},
        ticklabel style={font=\scriptsize},
        height=\plotheighthalf, width=\plotwidth,
        ]

    \addplot[mark=none, thick, color=darkgray100] 
    table {%
        0 0
        22 0
    };

    \addplot[mark=none, thick, color=red] table [x=time, y=drivewheel, col sep=comma] {figures/hardwareresults/wheelbot_hardware_turbo_initial.csv};

    \addplot[mark=none, thick, color=blue] table [x=time, y=drivewheel, col sep=comma] {figures/hardwareresults/wheelbot_hardware_turbo_best.csv};

\end{axis}
\node[above,font=\bfseries,xshift=3mm] at (current bounding box.north) {Mini Wheelbot};
\end{tikzpicture}%
        \vspace{-2em}
      \end{subfigure}
    \caption{
    Hardware results: Improvement from initial (\textcolor{red}{red}) to the optimized policy (\textcolor{blue}{blue}) is illustrated with closed-loop trajectories: on the cartpole, shorter swing-up and zero cart position (top right);
    on the Mini Wheelbot, reduced yaw overshoot and driving wheel action (bottom right).
    }
    \label{fig:hardwareresults:closedloop}
    \vspace{-0.5em}
\end{figure}
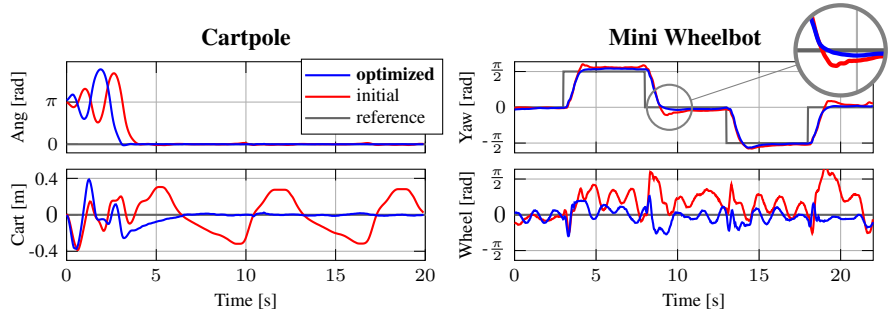

The hardware results are summarized in Fig.~\ref{fig:hardwareresults:improvement} and~\ref{fig:hardwareresults:closedloop}.
Compared to the initial guess, BO improves within 20 experiments (10-15 min of hardware interaction), which is consistent across multiple random seeds (Fig.~\ref{fig:hardwareresults:improvement}).
Qualitatively, for the pendulum, optimized parameters reduce the time required for the swing-up and drive the cart to the center of the rail during stabilization instead of oscillating around the center as depicted in Fig~\ref{fig:hardwareresults:closedloop}~(top).
On the reaction wheel unicycle robot, we can observe in Fig~\ref{fig:hardwareresults:closedloop}~(bottom) that the optimized policy reduces oscillations of the driving wheel (\ie less driving back and forth during maneuvers) and minimizes the overshoots during the yaw step response.
This improvement in qualitative performance for both systems is also clearly visible in the video of our experiments \videolink.

\section{Conclusion}
\label{sec:conclusion}
In this paper, we proposed a method for automatically fine-tuning an AMPC.
This eliminates the need for iteratively synthesizing datasets and retraining NN controllers -- a major drawback of classic AMPC in practice.
With our method, a single NN controller imitating a nominal MPC is sufficient, while the proposed automatic tuning adapts the AMPC to the actual hardware.
We achieve this by fine-tuning parameter-adaptive AMPCs~\cite{hose2024parameter} to optimal task- and hardware-specific performance with local BO from only a few hardware experiments.
We demonstrate the effectiveness of our method on two challenging, nonlinear, and unstable control tasks in simulation and hardware experiments:
a cartpole swing-up and balancing task, and a reaction wheel unicycle robot balancing and yaw control task.
In both setups, the neural network controller runs on embedded processors and is evaluated within milliseconds, which would not be possible with the classic optimization-based MPC we imitate.
BO consistently improves the initial parameters within 20 hardware interactions on both tasks.
We believe this combination of parameter-adaptive AMPC and automatic fine-tuning via BO has the potential to make AMPC a practical tool for a wide range of real-world control applications.

\fakepar{Limitations and Future Work}
We see three main limitations of the proposed method.
First, the used AMPC scheme can only adapt to parameter changes within a local region around the nominal parameters as it relies on the sensitivities of the MPC problem.
However, we empirically show, that they are sufficient to locally adapt policies and -- to some extent -- transfer to different instances from the same class of systems.
However, the sensitivities may not be accurate enough for vastly different systems or control objectives to achieve satisfactory performance.
Second, we only evaluated our method on a small and medium-sized system, which is good empirical indication that the method scales well.
However, it is unclear, how to scale the AMPC synthesis to very high dimensional states (\ie environments with hundreds of states or end-to-end learning from image data).
Lastly, the proposed method considers time-invariant parameters.
This might be an oversimplification in applications where parameters change over time, for example, due to wear and tear. In future work, we will tackle this last issue by investigating the usage of time-varying or event-triggered BO schemes~\cite{bogunovic2016time,brunzema2022event}.

\begin{credits}
\subsubsection{\ackname}
This work is funded in part by the German Research Foundation (DFG) – RTG 2236/2 (UnRAVeL), the DFG priority program 1914 (grant TR 1433/1-2), and the Robotics Institute Germany, funded by BMFTR grant 16ME0997K.
Simulations were performed with computing resources granted by RWTH Aachen University under project RWTH1570.
\end{credits}

\appendix
\label{sec:appendix}

\section*{Appendix 1: Bayesian Optimization Hyperparameters}
\label{app:hyperparameters}
\addcontentsline{toc}{section}{Appendix 1: Bayesian Optimization Hyperparameters}

In the following, in Table~\ref{tab:hypersbo}, we list all the hyperparameters for the reward functions of the cartpole and Mini Wheelbot, as well as the hyperparameters for TuRBO \cite{eriksson2019scalable}, to reproduce the results in Sec.~\ref{sec:results}.
We only list the hyperparameters from TuRBO that differ from the default values used in the corresponding paper as well as in the TuRBO implementation in BoTorch~\cite{balandat2020botorch}.

\sisetup{parse-numbers = false}
\begin{table}[h!]
\centering
\caption{Hyperparameters used in the Cartpole and Mini Wheelbot experiments.}\label{tab:hypersbo}
\begin{minipage}[t]{0.48\linewidth}
\centering
\subcaption{Cartpole experiments.}
\begin{tabular}{l c c c}
\toprule
\textbf{Param} & $\bar{s}\ulm{pos}$ & $w\ulm{pos}$ & TuRBO~$L\ulm{initial}$ \\
\midrule
\textbf{Value} & $\SI{0.39}{\meter}$ & $\SI{\frac{5}{0.39}}{\per\meter\squared}$ & $0.4$ \\
\bottomrule
\end{tabular}
\end{minipage}
\hspace{0.02\linewidth}
\begin{minipage}[t]{0.48\linewidth}
\centering
\subcaption{Mini Wheelbot experiments.}
\begin{tabular}{l c c c}
\toprule
\textbf{Param} & $w\ulm{yrp}$ & $w\ulm{wheel}$ & TuRBO~$\tau\ulm{fail}$ \\
\midrule
\textbf{Value} & $[1, 0.001, 0.01]^\top$ & $0.1$ & $3$ \\
\bottomrule
\end{tabular}
\end{minipage}
\end{table}

\fakepar{Number of Simulation Experiments}
Fig.~\ref{fig:simulationresults} results are for cartpole and Mini Wheelbot simulations with~\num{100} random systems each.

\fakepar{Number of Hardware Experiments}
Fig.~\ref{fig:hardwareresults:improvement} results on cartpole and Mini Wheelbot are for~\num{5} random seeds for TuRBO and~\num{5} random seeds for Sobol sampling each.

\section*{Appendix 2: Parameter Bounds}
\addcontentsline{toc}{section}{Appendix 2: Parameter Bounds}\label{app:parambounds}

Below are the parameter bounds used to synthesize random systems around the nominal parameters for the simulation results in Sec.~\ref{sec:results} for both cartpole and the Mini Wheelbot.

\sisetup{parse-numbers = false}
\begin{table}[h!]
\caption{Parameter bounds around the nominal parameters $\paramnom$}\label{tab:param_cartpole}
\centering
\subcaption{Cartpole}
\begin{tabular}{l c c c c c}
\toprule
\textbf{Param} & $m_\text{add}$ &  $M$ &  $C_1$ &  $C_2$ &  $C_3$\\
\midrule
Upper & $\SI{0.016}{\kilo\gram}$ &  $\SI{0.4}{\kilo\gram}$ & $\SI{2}{\newton\second\per\meter}$ & $\SI{0.4}{\newton\per\volt}$ & $\SI{0.008}{\newton\meter\second\per\radian}$\\
Lower & $\SI{-0.016}{\kilo\gram}$ &  $\SI{-0.4}{\kilo\gram}$ & $\SI{-2}{\newton\second\per\meter}$ & $\SI{-0.4}{\newton\per\volt}$ & $\SI{-0.008}{\newton\meter\second\per\radian}$\\
\bottomrule
\end{tabular}
\subcaption{Mini Wheelbot}
\centering
\resizebox{\textwidth}{!}{
\begin{tabular}{l c c c c c c c c c}
\toprule
\textbf{Param} & $m_\text{B}$ & $m_\text{W,R}$ & $I_{\text{B}, \{x, y, z\}}$ & $I_{\text{W,R}, \{y,z\}}$ & $I_{\text{W,R} \{x\}}$ & $r_\text{W,R}$ & $l_\text{WB}$ & $\mu_1$ & $\mu_2$ \\
\midrule
Upper & $\SI{0.1}{\kilo\gram}$ & $\SI{0.05}{\kilo\gram}$ & $\SI{10^{-4}}{\kilo\gram\meter\squared}$ & $\SI{20 \cdot 10^{-6}}{\kilo\gram\meter\squared}$ & $\SI{50 \cdot 10^{-6}}{\kilo\gram\meter\squared}$ & $\SI{0.005}{\meter}$ & $\SI{0.005}{\meter}$ & $0.01$ & $50$ \\
Lower & $\SI{-0.1}{\kilo\gram}$ & $\SI{-0.05}{\kilo\gram}$ & $\SI{-10^{-4}}{\kilo\gram\meter\squared}$ & $\SI{-20 \cdot 10^{-6}}{\kilo\gram\meter\squared}$ & $\SI{-50 \cdot 10^{-6}}{\kilo\gram\meter\squared}$ & $\SI{-0.005}{\meter}$ & $\SI{-0.005}{\meter}$ & $-0.01$ & $-50$ \\
\bottomrule
\end{tabular} }
\end{table}

\bibliographystyle{splncs04}
\bibliography{references}  %

\end{document}